\documentclass[runningheads]{llncs}
\usepackage{graphicx}
\begin{document}
\title{Ethics of Artificial Intelligence Demarcations}
%
%
\author{Anders Braarud Hanssen \and
Stefano Nichele}
\authorrunning{Hanssen and Nichele}

\institute{Oslo Metropolitan University, AI Lab, 
\email{\{anbh, stenic\}@oslomet.no}\\
}
\maketitle              
\begin{abstract}
In this paper, we present a set of key demarcations, particularly important when discussing ethical and societal issues of current AI research and applications. Properly distinguishing issues and concerns related to Artificial General Intelligence and weak AI, between symbolic and connectionist AI, AI methods, data and applications are prerequisites for an informed debate. Such demarcations would not only facilitate much-needed discussions on ethics on current AI technologies and research. In addition, sufficiently establishing such demarcations would also enhance knowledge-sharing and support rigor in interdisciplinary research between technical and social sciences. 

\end{abstract}
\section{Introduction}

The original goal of Artificial Intelligence (AI) research was to create an artificial (electronic) brain. This idea was explored in the seminal work by McCullock and Pitts \cite{mcculloch1943logical}, where they proposed a network of simplified abstract versions of biological neurons. The goal of creating a full artificial brain with the same degree of intelligence of a human brain is still an open challenge. From the idea of a brain capable of general (human) intelligence, the interest of the AI community quickly moved towards simplified (narrow) versions of artificial intelligence, to solve specific tasks.

The state-of-the-art in (narrow) AI was described by D. Waltz on the Scientific American back in 1982 \cite{waltz1982} as \textit{"Computer programs that not only play games but also process visual information, learn from experience and understand some natural language"}. In addition, he added that \textit{"The most challenging task is simulating common sense"}. The current state-of-the-art in AI has not changed radically from Waltz's definition. Today the most compelling and less understood aspect is still the simulation of common sense, i.e., reasoning and cognition. The scaling of computational resources has allowed advances in playing computer games, computer vision, and natural language processing, pretty much with the same methods used in the '80s. While the initial AI inspiration was the human brain, in the meanwhile several methods to simulate intelligence without neural-based systems emerged, e.g., symbolic AI. Such methods had a certain degree of success thanks to the less need for computational resources. The recent availability of massive computational resources has allowed scaling of neural systems with results that surpassed non-neural systems in most application domains.

In December 2018, the European Commission's High-Level Expert Group on Artificial Intelligence has proposed the following updated definition of AI \cite{AI2018EU}:

\textit{"Artificial Intelligence (AI) refers to systems designed by humans that, given a complex goal, act in the physical of digital world by perceiving their environment, interpreting the collected structured or unstructured data, reasoning on the knowledge derived from this data and deciding the best action(s) to take (according to pre-defined parameters) to achieve the given goal. AI systems can also be designed to learn and adapt their behaviour by analyzing how the environment is affected by their previous actions. As a scientific discipline, AI includes several approaches and techniques, such as machine learning (of which deep learning and reinforcement learning are specific examples), machine reasoning (which includes planning, scheduling, knowledge representation and reasoning, search, and optimization), and robotics (which includes control, perception, sensors and actuators, as well as the integration of all other techniques into cyber-physical systems)."}

The current understanding of AI ethics is rather vague, due to the broad definitions of AI used in the literature, and do not necessarily reflect the aspects and demarcations within the research community, the algorithms and methods, the computing substrates \cite{konkoli2018philosophy}, and the target applications.

In the remainder of this paper, we will outline and discuss some important AI demarcations which have strong implications for the ethical aspects and possible reflections to address key issues in research on societal impacts such as Responsible Research and Innovation (RRI).

\section{AI Demarcations}

\subsection{Weak AI vs AGI}

The first and perhaps the most well known AI demarcation is the one between Weak AI (also known as Narrow AI, Applied AI) and Artificial General Intelligence (also called Strong AI or Full AI). While Weak AI aims at making a machine learn to solve a specific task, AGI targets machines that can learn and perform any intellectual task. This implies that AGI has the ability to "learn to learn", as well as the ability of problem-solving, reasoning, modelling and planning. G. Marcus and Y. LeCun, two prominent AI researchers, while disagreeing in many aspects of the future of AI, agree on a list of seven points \cite{Marcus2017}: 

\begin{itemize}
  \item AI is still in its infancy
  \item Machine learning is fundamentally necessary for reaching strong AI
  \item Deep learning is a powerful technique for machine learning
  \item Deep learning is not sufficient on its own for cognition
  \item Model-free / Reinforcement learning is not the answer, either
  \item AI systems still need better internal forward models
  \item Commonsense reasoning remains fundamentally unsolved
\end{itemize}

It is evident that the demarcation between weak AI (all AI today and in the near future) and AGI implies that all current methods do not incorporate any form of commonsense reasoning, and the most used method of deep learning is not sufficient for a truly cognitive system. In addition, there is no current understanding or scientific theory on how commonsense reasoning could be achieved. 

\subsection{Symbolic AI vs Connectionist AI}

\begin{figure}[t]
\includegraphics[width=\textwidth]{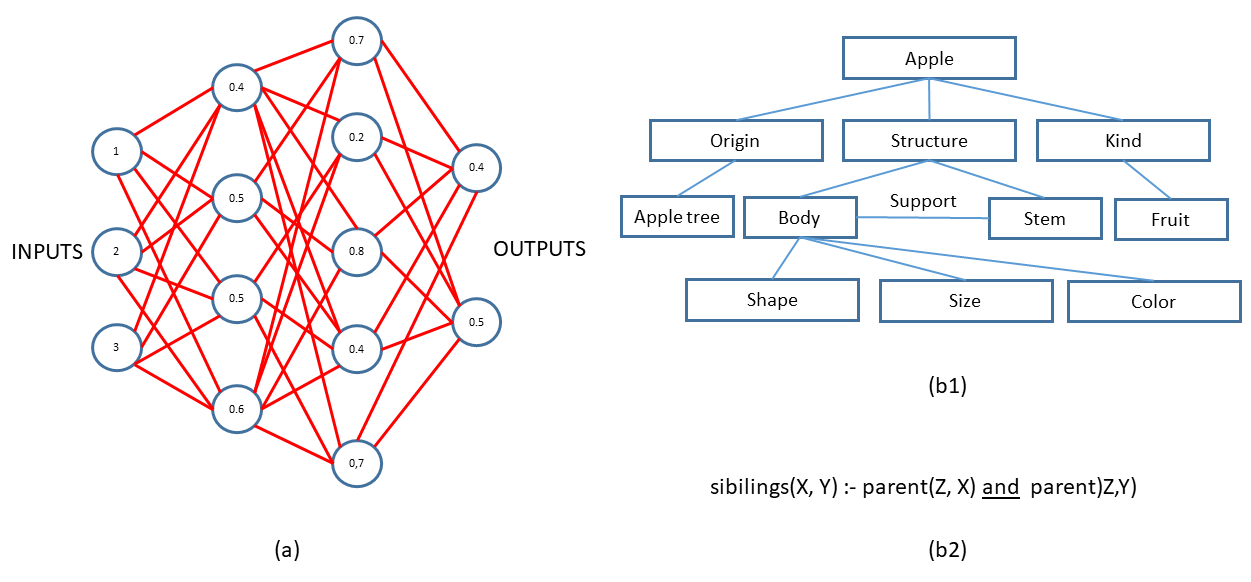}
\centering
\caption{(a) connectionist representation where information is represented by synapses (red lines) between neurons (blue nodes). (b) two examples of symbolic representations, b1 with a tree representation and b2 with logic expression.}
\label{fig:img1}
\end{figure}

Another important demarcation for AI systems is represented by the way information and relations are represented and encoded. In symbolic AI (also called algorithmic AI), knowledge is encoded in a symbolic form, together with rules to manipulate symbols and their relations. While symbol representation and manipulation makes it possible for a more rigorous study and explanation of weak AI systems, there is no evidence that the human brain is programmed as a symbolic machine. On the other hand, connectionist AI refers to  a large network of units (neurons) that are interconnected together and encode/process information in a distributed way. While such models are more biologically plausible, they are typically data and compute hungry. Examples of symbolic and connectionist AI representations are depicted in Figure \ref{fig:img1}.

\subsection{AI method vs Data}

One demarcation that is often confused, especially in the context of AI bias, is the dataset used to train the AI model vs. the learning algorithm used to train the AI model (Note: the result of the training process using a specific dataset is an actual trained model, see next subsection). The fact that a trained model is biased is a feature of the AI model and not a bug. In fact, if one wants to model a real-world system, the actual real-world model may be biased. The training algorithm is transparent to bias and therefore should not be attributed for the AI system being biased. If the intention of the AI model is to be unbiased, then the used dataset (the sole source of bias) should be corrected. One example of training algorithm for neural networks is backpropagation. Backpropagation involves mathematical operations such as calculating the derivative of the squared error function with respect to the weights of the network. This type of mathematical operations does not allow for algorithmic bias.

\subsection{AI method vs application}

The actual trained AI model, and therefore the application in which the AI is used, is not to be confused with the AI algorithm or method used for training. This demarcation is very important as restrictions have to be considered at the application level rather than at the AI method and algorithmic level (i.e., the method may be the same in very different domains, and obviously with very different sets of data). An example: regulating databases. It is the wrong level of abstractions. What is regulated is the use cases of databases (e.g., credit card companies, or insurance companies). We regulate lawyers, not Word. We regulate financial companies, not Excel. We do not regulate steel companies, we regulate guns. And we do not ask steel companies to regulate guns. AI is not an application, it is a general set of building blocks. 

\subsection{AI vs humans}

Would regulators approve and consider ethically acceptable human-driven cars if they were invented today? Probably not. They are very dangerous by today's ethical standards. One important demarcation is human intelligence vs. artificial intelligence. Many of the issues with artificial intelligence are also present in human intelligence, e.g., black box. Can our intelligence be inspected when we drive a car? Is human intelligence open-source? Is the intelligence architecture known? Is the data used to train us for driving biased? Yes indeed. As we are given different datasets when we learn to drive. Is human intelligence deterministic? Can human intelligence be evaluated under different environmental conditions or noise? Are experiments repeatable?
Those are all relevant questions to better understand the demarcation between human intelligence and AI.

\subsection{Embodied AI}

Features of the human cognition are shaped by aspects of the body (beyond the brain) \cite{pfeifer2006body}. Intelligence and cognition include high-level mental constructs (concepts and categories) and human performance on various cognitive tasks (reasoning / judgment), as a result of embodiment. Aspects of the body that shape cognition include the motor system, the perceptual systems, as well as the body interactions with the environment. It is therefore expected that artificial general intelligence requires embodied agents living in an environment. One may argue that weak AI lacks often embodiment and a reactive environment.

\subsection{Compelling questions}

Through the description of the AI demarcations above, a list of relevant compelling questions for AI ethics emerged, and is listed below:

\begin{itemize}
\item What do we consider artificial intelligence?
\item Are intelligent machines considered living machines?
\item Can we demonstrate the emergence of intelligence and mind in an artificial living system?
\item What ethical principles should be established for artificial general intelligence and weak artificial intelligence?
\item What role do societal and ethical perspectives play in understanding the difference between human and machine intelligence?
\end{itemize}

\section{Analysis}
Based on the above synopsis of demarcations, we turn to how ethical and societal considerations are addressed within the generic field of AI. Current ethical and social science issues in and around artificial intelligence may benefit from a more rigorous articulation of demarcations within AI research. However, an overview of such issues would first benefit from situating applied ethics in new and emerging technologies. 

Applied ethics as a discipline could be understood as relating to various practical applications of moral thought and principles and has a longstanding role within such fields as medicine, law and within various processions. In recent years, a range of approaches within ethics has addressed applications and implications of various ethical concerns within AI and machine learning \cite{luxton2015artificial}, \cite{etzioni2017incorporating}. Nevertheless, scholars have argued that merely addressing specific technical and ethical concerns in isolation may not be the most viable approach to the legitimacy of the future of AI research and applications \cite{cath2018governing}. The humanities and social sciences have gained validity when facing the many uncertainties related to societal impacts of new and emerging technologies. Arguably, AI research poses unprecedented societal and ethical questions both to the nature of such research and its outcomes. Thus, specific ethical questions and implications could also be seen in a broader context. Among such broader considerations are the importance of stakeholder involvement, transparency and accountability. Such issues engage considerations beyond the discipline of applied AI ethics and involve questions of governance and policy. As a consequence, AI research benefits from research addressing these concerns in particular. But what kinds of research incorporate these broader concerns and how does such research sufficiently incorporate the necessary rigor related to technical demarcations within various sub-fields of AI?

Within new and emerging technologies, ethical and societal considerations gained prominence after the surge in genomic research in the U.S through the Human Genome Project (HGP) under the label of ethical, legal and social implications (ELSI)\cite{fisher2005lessons}  and later through its European counterpart ELSA. ELSI Research was seen as a necessary component of addressing potential social and ethical implications of the vast uncertainties related to genomic research, particularly through its commercialization. These avenues of research have in recent decades been applied to new and emerging technology areas such as nanotechnology \cite{hullmann2008european}, synthetic biology \cite{coenen2009ethics} and various ICT research \cite{nydal2015ethics}. After 2010 these research areas, through increased awareness of policy considerations combined through the term responsible research and innovation (RRI), which soon was adopted by the European Union´s Framework Programmes. From both a research and policy perspective RRI emphasized the need to take the societal, ethical and environmental impacts of emerging science and technology into account. Simultaneously RRI has emphasized the need to align research and innovation with societal challenges. Nevertheless, research on applied ethics on AI and also RRI-informed research on AI is still in its infancy and is fraught with many shortcomings. Among these are establishing necessary demarcations and distinctions that are both epistemic and normative in nature. In fact, very few examples exist in current RRI-literature where a clarification between key concepts and issues in both research and applications are undertaken in a systematic manner. Such clarifications and demarcations would inform the trajectory of various discussions around ethics and societal impacts.  Arguably they would also contribute to establishing a better understanding and learning outcomes between AI researchers, ethicists and social scientists. To further illustrate the importance of such demarcations in the context of RRI- or other forms of social science-based research, a few key examples will be presented in the following paragraphs.

Sufficiently distinguishing between weak AI and AGI underscores the need to separate broad debates on AGI from timely and necessary reflection on societal embedding of weak AI. Although compelling, AGI debates are marked by both dystopian and utopian narratives and based on probability and hype \cite{cave2018portrayals}. Moreover, obvious knowledge gaps in the current research frontier seem to under-emphasize the limitations of the current understanding of common sense reasoning and cognition in humans. Such limitations are currently making the realization of superintelligent AGI unforeseeable \cite{wang2018conceptions} .  Nevertheless, the recurring worst-case scenarios and hype of AGI threaten the legitimacy of various weak AI applications and research in the general public. These debates may also overshadow the need to address pressing questions in relation to governance and regulation or areas where weak AI already is being implemented. Moreover, weak AI-based research frequently lacks the presence of integrated social science and ethical perspectives in their design. Such perspectives may contribute to a broader understanding of challenges within areas such as machine learning. Designed on the semblance of human learning, it draws from cultural and social structures and extrapolates from them. However, a better understanding of how algorithms build on such structures would also inform our understanding of what they cannot do, i.e such as present solutions for any scenario.

The nature of algorithmic design also shows that ethical and societal issues may be of very different natures with regard to current symbolic and connectionist AI and thus also provide very different ethical and societal scenarios.  Symbolic AI may have vulnerabilities related to the quality of the design and/or hidden bias embedded into the algorithm itself, i.e bias related to relationships or symbols within symbolic language such as representing 'nurse' as 'female'. Although easily correctable it shows that ethical considerations such as gender equality point back to tacit linguistic biases and cannot be seen in isolation. At the same time, symbolic AI yields greater transparency towards such  bias. Connectionist AI, as in deep neural networks, represents concerns more related to accountability and lack of transparency of what now seems to be 'black box-issues'. These may involve biases embedded in the data sets, such as societal, linguistic, cultural and heuristic biases that are embedded in data while at the same time present correlations that are context-sensitive, such as deep neural networks that may successfully predict sexual orientation by image analysis \cite{wang2018deep}. Further, by being 'data-hungry', connectionist AI seems vulnerable to error if data sets are not sufficiently substantial. Thus, in this regard ethical discourse around symbolic AI may yield results swiftly while in various connectionist AI context-of-use scenarios may be the most viable area of study. 

The demarcation between the trained (applied) AI model and the training algorithm should inform what forms of ethical considerations are addressed; considerations that may often be misplaced. Some scholars argue regulation and law should primarily focus on the use of the model while the training algorithm itself could be considered as merely a tool \cite{vayena2018machine}. Others argue that regulation and standardization are equally important in both \cite{goodman2017european}. Nevertheless, arguments that the training algorithms themselves are biased could be resolved by a proper demarcation between the training algorithm and application of the model. However, more research on the value-assumptions embedded in algorithmic training is needed, particularly in the discrepancy of the data embedded in the training algorithm and real-life scenarios \cite{dent2018ethical}. Beyond these demarcations, there are different ethical considerations to be accounted for in the role that certain data sets play in the model and how an AI-tool is applied to various decision-making situations. In particular, if the bias is unknown or unidentified before the model is implemented it may have downstream impacts. Thus, in a range of scenarios, considerations such as distributive justice and or privacy may engage concerns related to both the training algorithm and the applied model. The data used in the training algorithm and the context of which the trained model is applied may in different ways combine to produce a complex set of urgent ethical and societal considerations. Nevertheless, distinguishing between the algorithm itself and the application of the model may itself resolve a range of unnecessary discussions about AI bias.

The distinction and/or similarities between human and artificial intelligence in relation to autonomous systems points to a discussion about the role of 'ethical algorithms' that in many situations may be a misplaced concern. Albeit debates of whether the 'Trolley-problem'  poses a contrived and unrealistic dilemma between utilitarianist and deontological reasoning in real-life scenarios \cite{de2019doubting}, its resolution may reside elsewhere. Both humans and autonomous systems opaque reasoning may in decision-making situations pose unreasonable risks and uncertainties. However, as of yet, the unforeseen consequences of outsourcing legal agency and responsibility from humans to autonomous systems may potentially be a more considerable societal risk. The incommensurability of legal, ethical and scientific reasoning may here be a more pressing subject than making algorithms 'ethical' and should address such issues as the problem of legal accountability.  To what extent may we accept bias in humans while not in autonomous systems if we consider autonomous systems bias to be a liability? Some may argue that ethical considerations for humans and machines should be considered distinct and separate. Humans are by default prone to error while legally accountable. Machines, who we seek to error-correct all the time, may be equally imperfect while in particular scenarios present algorithmic decision-making that seems ethically 'superior' to human action.  While the question remains if we should base moral judgments on the outcomes or intention of an action, demarcating human and machine 'ethics' is a pressing concern. It would at least seem important to define what moral status human and machine action have when they are equally nontransparent. 
However opaque, human intelligence is a product of adaptation to the environment. This embodied aspect of intelligence may at least provide us with a demarcation between machine and human intelligence (embodied vs disembodied). Weak AI, by virtue of lacking embodiment, substantially differs in nature from human intelligence. It would thus follow that there should be other ethical considerations (i.e moral rights and obligations) for weak AI systems than embodied systems. 

\section{Conclusions}
For the purpose of our discussion, a substantial part of the current debate about AGI evolves around threats and promise based on speculation. However, such concerns are less pressing and bound by considerable uncertainties and unresolved scientific challenges to develop a fully cognitive system. We argue that a proper demarcation between AGI and weak AI would facilitate a more informed debate about pressing concerns related to challenges and opportunities. Such scenarios are worthy of discussion, not only for ethicists but for AI research institutions,  policy-makers and for society-at-large. Further, when ethical and societal impacts are discussed, a distinction between issues related to connectionist and symbolic AI is needed to be able to identify both vulnerabilities, risks and countermeasures. Similarly, in discussing AI bias, distinguishing between the AI method and the bias embedded into the data, the training algorithm itself and the trained model would resolve uncertainties and identify how and to what extent bias should play a role in training algorithms. Further, it would clarify where ethical and societal issues may most adequately be identified. Sufficiently clarifying the difference between human and machine ethics points back to the insufficient demarcation between human and machine reasoning which is often equally opaque. Such a clarification would inform the debate on how to proceed with developing a more feasible 'machine ethics' and 'ethical algorithms', and potentially point the attention to issues of legal accountability.

It has been the overarching objective of this paper to illustrate the need for more rigor in the discourse on ethical and societal impacts of AI research in relation to a lack of sufficiently demarcating between key features of AI methods and tools. By providing illustrations of key demarcations, we have also suggested to what extent this may inform research on ethical and societal issues, both on weak AI and AGI. Further, through establishing the relevance such demarcations in the context of ethical and societal impacts,  a range of ongoing discussions about AI could adopt them to provide a more nuanced and more elaborate dialogue across disciplinary boundaries. In particular, such a broader approach to ethical and societal impacts of AI should shift focus from isolated and narrow ethical questions to include governance and regulatory considerations and facilitate knowledge-sharing among stakeholders. However, approaches such as RRI may not successfully engage in knowledge-sharing with AI researchers if the aforementioned demarcations are not taken sufficiently into account.

%
%
%
\bibliographystyle{splncs04}
\bibliography{bib}

\end{document}